\documentclass[conference]{IEEEtran}

\usepackage{latexsym}

\usepackage{times}
\usepackage{amsfonts}
\usepackage{amsmath}
\usepackage{multirow}
\usepackage{times}
\usepackage{helvet}
\usepackage{courier}
\usepackage{graphicx}
\usepackage{multirow} 
\usepackage{xcolor}
\usepackage{subfigure}

\ifCLASSINFOpdf
\else
\fi
\begin{document}

\title{A Multi-Modal Chinese Poetry Generation Model}

\author{\IEEEauthorblockN{Dayiheng Liu}
\IEEEauthorblockA{Machine Intelligence Laboratory \\ College of Computer Science\\
Sichuan University \\ Chengdu 610065, P. R. China \\
Email: losinuris@gmail.com}
\and
\IEEEauthorblockN{Quan Guo}
\IEEEauthorblockA{Machine Intelligence Laboratory \\ College of Computer Science\\
Sichuan University \\ Chengdu 610065, P. R. China \\
Email: guoquanscu@gmail.com}
\and
\IEEEauthorblockN{Wubo Li\\ and Jiancheng Lv}
\IEEEauthorblockA{Machine Intelligence Laboratory \\ College of Computer Science\\
Sichuan University \\ Chengdu 610065, P. R. China \\ Email: lvjiancheng@scu.edu.cn}}

\maketitle

\begin{abstract}
Recent studies in sequence-to-sequence learning demonstrate that RNN encoder-decoder structure can successfully generate Chinese poetry. However, existing methods can only generate poetry with a given first line or user's intent theme. In this paper, we proposed a three-stage multi-modal Chinese poetry generation approach. Given a picture, the first line, the title and the other lines of the poem are successively generated in three stages. According to the characteristics of Chinese poems, we propose a hierarchy-attention seq2seq model which can effectively capture character, phrase, and sentence information between contexts and improve the symmetry delivered in poems. In addition, the Latent Dirichlet allocation (LDA) model is utilized for title generation and improve the relevance of the whole poem and the title. Compared with strong baseline, the experimental results demonstrate the effectiveness of our approach, using machine evaluations as well as human judgments.
\end{abstract}

\IEEEpeerreviewmaketitle

\section{Introduction}
China is known as ``the kingdom of poetry". This is not only because of the long history of Chinese poetry, but also the number of poets and works, which has a special and significant place in Chinese social life and cultural development.

\begin{figure}[!t] 
   \centering
   \includegraphics[width=3.5in]{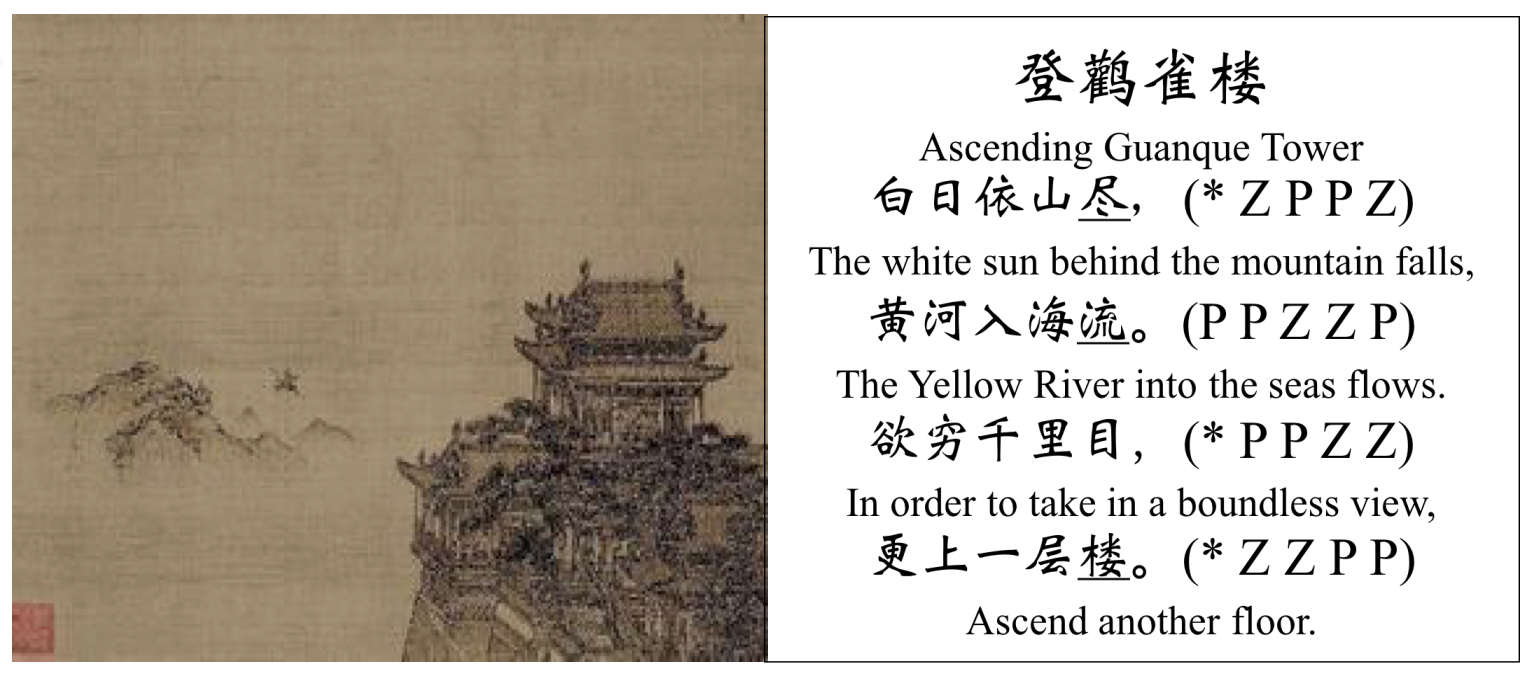}
   \caption{An example of 5-character \textit{quatrains}. The poet wrote the poem moved by the sight on the left of the figure. The rhyming characters are shown in underline. The tone of each character is shown at the end of each line (within parentheses); P and Z are short-hands for Ping and Ze tones respectively; * indicates that the tone is not fixed and can be either. }
   \label{fig:figure1}
\end{figure}

Poetry is the carrier of language, the most original and the most authentic art. It is engraved with human reason and emotion, wise and thought, imagination and shouting, rough and smooth. There are several writing formats for Chinese Tang poetry, among which \textit{quatrain} is perhaps the best-known one which requires strict rules including words, rhyming, tone and antithesis, we illustrate an example of famous quatrains in Figure \ref{fig:figure1}.

1) Words. The \textit{quatrain} consists of 4 lines of sentences, and the length of each line is fixed to 5 or 7 characters.

2) Rhyming. The syllables of Chinese characters are composed of initials and finals. Rhyming words must have same finals. Poetry pays attention to the beauty of melody and rhythm. Therefore, the Chinese poetry must rhyme. Rhyme means to put the same rhyme in the same place, usually at the end of a sentence (the underlined characters in Figure \ref{fig:figure1}).

3) Tone. This is the character of Chinese. The height, the rise and fall and the length of speech form the tones of Chinese. The four ancient sounds were: level tone, rising tone, falling tone, and entering tone. The relationship between the four tones and the rhyme is very close. The words in different tones usually can't rhyme. Poets divided the four voices into two broad categories: Ping (level tone) or Ze (down-ward tone). The Ping and Ze alternates in the verses according to certain rules, so that the tone is diversified rather than monotonous.

Writing a good poem is difficult and poets need to master profound literary skills. It is hard to master such skills for ordinary people. In recent years, automatic Chinese poetry generation has made great progress. There are several different kinds of approaches to generate poems. One of the most promising approaches is taking the generation of Chinese poem lines as a sequence-to-sequence learning problem \cite{sutskever2014sequence,cho2014properties}. The RNN Encoder-Decoder model with attention mechanism \cite{bahdanau2014neural,luong2015effective} is employed to generate Chinese poems. It has been shown that this sequence-to-sequence (seq2seq) neural model can successfully generate Chinese poems \cite{Wang2016Can,Yi2017Generating}. 

However, there are still some defects in these existing approaches: (i) These methods can only generate a poem with a given first line or the user's intention, and often generate non-thematic poetry or the theme of the whole poetry is not consistent with the theme of the user's intention. (ii) Some properties of quatrains such as symmetry are not considered. (iii) They cannot generate titles for the generated poems.

To address these issues, we propose a multi-modal three-stage approach to generate the Chinese \textit{quatrain} and its relevant title with a given picture or a theme: 1) We first obtain a theme related phrase from an external knowledge base called \textit{ShiXueHanYing}. If the input is a picture, it is mapped into a specific theme with a GoogleNet \cite{szegedy2015going,szegedy2015rethinking} called \textit{image recognition module} which is fine-tuned on our manually build dataset. To enhance the relevance of the poem and the theme, the Backward and Forward Language Model \cite{Mou2016Backward,Mou2016Sequence} (B/F-LM) with GRU cell \cite{Cho2014Learning} are employed to generate the first line of the poem which explicitly incorporates the theme related phrase. 2) After the first line generation, we utilize an LDA model to find a suitable theme related phrase from \textit{ShiXueHanYing} as the title after first line generation. This title is going to guide other lines' generation to make the whole poem more relevant with the title. 3) We propose a hierarchy-attention seq2seq model (\textit{HieAS2S}) to generate the remaining poem line by line. This model can effectively capture character, phrase, and sentence information between contexts and improve the symmetry delivered in poems. 

For machine evaluation, we modify the BLEU evaluation which is used in \cite{Wang2016Can,zhang2014chinese} to find more suitable references for evaluation. Furthermore, we regard whether the generated poems satisfy the rhyme and tone as an additional evaluation. Our experimental studies indicate that the proposed HieAS2S model outperforms several variants of seq2seq model. The proposed three stage method performs better than strong baselines in both machine and human evaluation. In addition, our title generation method performs completely well when compared with standard seq2seq model with attention mechanism. Particularly, we develop a web application for users to use and evaluate our approach. Most users give satisfactory evaluations.

\begin{figure*}[t] 
   \centering
   \includegraphics[width=7in]{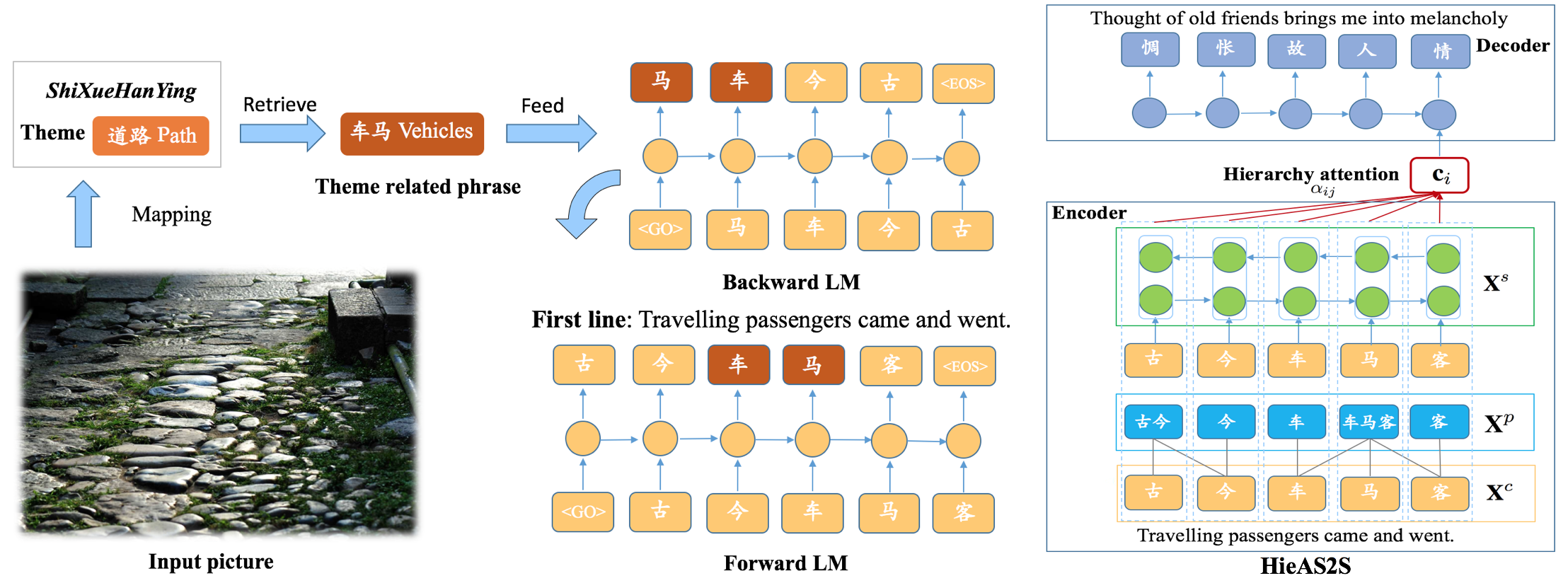} 
   \caption{The framework of our multi-modal three-stage generation approach (best viewed in color). The part of the title generation is omitted. Given a picture example, the \textit{image recognition module} first map it into a \textit{ShiXueHanYing} theme `path'. Then the theme related phrase `CheMa' is randomly picked. We reverse and input it to the backward LM to generate the first half of the first line. This result is fed to the forward LM to generate the whole first line ``Travelling passengers came and went''. The right part of the figure shows the architecture of the \textit{HieAS2S} model. Given previous generated lines, we extract their character level, phrase level and sentence level features as the hierarchy attention memories to calculate the attended context vectors. The next line ``Thought of old friends brings me into melancholy.'' is generated by the RNN decoder inputting the attended context vectors.}
   \label{fig:figure2}
\end{figure*}

\section{Related Work}
As poetry is one of the most significant and popular literature all over the world, the topic of poem generation has attracted lots of researchers over the past decades. There are several different kinds of approaches to generate poem. The first kind of approach is based on templates and rules. For instance, leveraging semantic and grammar templates \cite{oliveira2012poetryme}, based on WordNet \cite{fellbaum1998wordnet} and parts of speech \cite{agirrezabal2013pos}, word association norms \cite{netzer2009gaiku}, genetic algorithms \cite{manurung2012using, manurung2004evolutionary}, and text summarization \cite{yan2013poet}. In these papers, templates are employed to construct poems according to a set of constraints such as rhyme, meter, stress and word frequency. The second kind of method is based on statistical machine translation (SMT). For Chinese couplets\footnote{A pair of lines of poetry which adhere to certain rules.} generation, \cite{jiang2008generating} translate the first line to the second line by using a phrase-based SMT approach. And \cite{he2012generating} expand this method to generate four-line Chinese quatrains. 

With the development of deep learning on natural language generation, neural networks have been applied on poetry generation. \cite{zhang2014chinese} first presents a model for Chinese poetry generation based on recurrent neural networks. Given some input keywords, they use a character-based RNN language model \cite{mikolov2010recurrent} to generate the first line, and then the other lines are generated sequentially by a variant RNN. \cite{wangQixin2016} use Long Short-term Memory (LSTM) based seq2seq model with attention mechanism to generate Song Iambics. Then \cite{Wang2016Can} extend this model to generate Chinese quatrains. Furthermore, \cite{yan2016poet} propose a RNN based model with attention mechanism and polishing schema to generate Chinese poems and Chinese couplets \cite{yanchinese}. To ensure that the generated poem is coherent and semantically consistent with the users' intents, \cite{Wang2016Planning} propose a two-stage poetry generating method. They first plan the sub-topics of the poem and then generate each line using a modified RNN encoder-decoder model. \cite{Yi2017Generating} take the generation of poem lines as a sequence-to-sequence learning problem. They build three poem line generation blocks based on RNN Encoder-Decoder (word-to-line, line-to-line and context-to-line) to generate a whole quatrain. More recently, \cite{Zhang2017Flexible} propose a simple memory-augmented neural model to generate innovative poems. \cite{Yang2017Generating} employ Conditional Variational AutoEncoder (CVAE) \cite{Fabius2014Variational,Yan2015Attribute2Image,Bowman2015Generating} for Chinese poetry generation.

Our approach is closely related to the works of deep learning mentioned above. However, several important differences make our approach novel: 1) Above mentioned methods cannot generate the titles for the generated poems. We propose a three-stage approach to generate Chinese \textit{quatrains}. This approach is able to generate high-quality \textit{quatrains} with relevant titles. 2) We introduce a multi-modal way for Chinese \textit{quatrains} generation. We extend the generation way to support poetry generation through pictures. 3) According to the characteristics of Chinese poems, we first incorporate phrase feature to poetry generation model and propose the \textit{HieAS2S} model. This model can effectively capture character, phrase, and sentence information between contexts and improve the symmetry delivered in poems. 

\section{Approaches}
In this section, we introduce the algorithm of our approach step by step, including: 1) \textbf{generating the first line} which explicitly incorporates the theme related phrase, 2) \textbf{title generation} with LDA model, 3) \textbf{other lines generation} with \textit{HieAS2S} model. The framework of our generation approach is shown in Figure \ref{fig:figure2}.

\subsection{First Line Generation}
Existing approaches usually generate poems by given users' intent themes. We extend this way to support the generation from pictures. The algorithm of first line generation is shown in the left side of the Figure \ref{fig:figure2}. Given a picture, we first map it into a specific \textit{ShiXueHanYing} theme with \textit{image recognition module}. \textit{ShiXueHanYing} is a poetic phrase taxonomy organized by Wenwei Liu in 1735 which consists of 1,016 manually picked themes. Each theme contains dozens of related phrases. There are more than $ 40,000 $ phrases in total and the length of each term is between $ 2 $ and $ 5 $ Chinese characters. 

For \textit{image recognition module}, we retrain the final layer of a GoogleNet which has been fully trained on ImageNet \cite{Deng2009ImageNet} before. The dataset we use for this retraining is a manually build image dataset according to the themes of \textit{ShiXueHanYing}. Because there are many fine-grained or abstract themes which are difficult for classification, we manually cluster and filter the themes into 40 classes. Then we build a large picture dataset labeled with the theme called \textit{PoetryImage} which has more than 40,000 pictures in total and about 1000 pictures for each class as training set and 100 for each as test set. The top-1 error rates of our \textit{image recognition module} in the test set is 7.8\% while the top-3 error rates is 4.7\%.

After mapping the given picture into a theme, we retrieve all related phrases and randomly pick one. Then we employ the B/F-LM to generate the first line which explicitly incorporates this theme related phrases. As shown in Figure \ref{fig:figure2}, the B/F-LM consists of a backward RNN language model and forward RNN language model with GRU cell. Since we know a \textit{prior} theme related phrase should be appear in the sentence, we reverse the theme related phrase and start with it to generate the backward sequence using the backward RNN. Then we feed the result to forward RNN to generate the whole line. 

\subsection{Title Generation}
Although the seq2seq model with attention mechanism has achieved good results on abstractive summarization \cite{Rush2015A,Chopra2016Abstractive}, we find it performs poorly for Chinese poetry title generation. Through our analysis, there are two main reasons: 

1) The titles in the training datasets contain a lot of \textit{noise}. For example, many poets are used to taking the titles according to their surroundings while writing poems. These titles usually contain some specific names of persons or landscapes. 

2) Generating the title end-to-end from a poem which is already formed by highly concise language is a difficult task. 

The model can be easily overfitting and generate some unsuitable titles which may contain some unrelated person names and place names. Because of the first aforementioned reason, we also rule out matching-based methods to find a title of a human-written poem from the corpus for a generated poem. Finally, we find a better method to indirectly generate the title. Instead of generating the poetry title after the whole poem has been generated, we employ an LDA model to find a suitable phrase from \textit{ShiXueHanYing} as the title after the first line generated. Then this title is going to guide other lines' generation to make the whole poem more relevant with it.

As topics have long been investigated as the significant latent aspects of terms, we use a large corpus includes Chinese poems, Song Iambics and ancient Chinese proses to train a 100-topic LDA model. After training, we obtain the probability distribution vector $\textbf{T}$ of phrase $ t_i $ belonging to each topic $ z_j $, which is
\begin{equation}	
	\textbf{T}(t_i) = [P(t_i | z_1), P(t_i | z_2), \cdots ,P(t_i | z_{100})].
\end{equation}
We define the relevance coefficient $ \phi $ of phrase $ t_i $ and $ t_j $ as follows:
\begin{equation}	
	\phi(\textbf{T}(t_j), \textbf{T}(t_j)) = 0.5 \times \frac{\textbf{T}(t_i) \cdot \textbf{T}(t_j)}{\Vert \textbf{T}(t_i)  \Vert \Vert \textbf{T}(t_i)  \Vert} + 0.5.
\end{equation}
After first line generated, the first line is segmented into several phrases $ S^1 =\{t'_1, \cdots, t'_k \} $. Then we select the most suitable phrase $ t^* $ which do not appear in the first line from all theme related phrases as the title:
\begin{equation}	
	t^* = arg\mathop{max}_{t \notin S^1}{\sum_{t'_k\in S^1}{\phi(\textbf{T}(t'_k), \textbf{T}(t))}}.
	\end{equation} 
Since not all phrases are suitable as titles, we use POS tagging to restrict what phrases of \textit{ShiXueHanYing} can be alternative titles in advance.

\subsection{The hierarchy-attention seq2seq model}
According to the characteristics of Chinese poetry such as symmetry, we propose a hierarchy-attention seq2seq model called \textit{HieAS2S} for other lines generation. Compared with the standard seq2seq attention model, this model can effectively capture the information of context at hierarchical scales, \textit{i.e}, character, phrase and sentence level.

After generating the first line and the title, the other lines are generated successively. Given previous $m$-1 generated lines $\{S^1, ..., S^{m-1}\}$, the \textit{HieAS2S} model models the probability of the $m$-th line $P(S^m | S^1,...,S^{m-1})$. For simplicity, we use $S^m_{1:t}$ to denote the first $t$ characters of $m$-th line. According to the probability theory, we have:
\begin{equation}
		 P(S^m_{1:T}) = \prod_{t=1}^T P(y_t | S^m_{1:t-1},S^1,...,S^{m-1}). 
\end{equation}
Here $y_t$ is the $t$-th character of the $m$-th line and $T$ is the length of sentence $S^m$.

\textbf{Hierarchy Memory}. The architecture of the \textit{HieAS2S} model is shown in the right of Figure \ref{fig:figure2}. Firstly, we introduce the encoder part. The one-hot character vectors of current generated lines are individually mapped into a $d$-dimensional vector space $\textbf{X}^c = [\textbf{x}^c_1, .., \textbf{x}^c_T] \in \mathbb{R}^{d \times T}$. We use pre-trained character embeddings which are trained on a large external corpus. Then a bi-directional RNN \cite{schuster1997bidirectional} with GRU cell converts these vectors into two sequences of d-dimensional vectors $\textbf{X}^s = [\textbf{x}^s_1, .., \textbf{x}^s_T] \in \mathbb{R}^{2d \times T}$ to capture sentence information. 

To consider the phrase information, similar to \cite{Lu2016Hierarchical}, we apply 1-D convolution with three different filter window sizes (unigram, bigram and trigram) on the character embedding vectors to obtain phrase features. At each location $t$, we compute the inner product of the character vectors with different window size filters:
\begin{equation}
		 \hat{\textbf{x}}^p_{s,t} = tanh(\textbf{W}^s \cdot \textbf{x}^c_{t:t+s-1}) \in \mathbb{R}^{d}, \qquad s \in \{1,2,3\} 
\end{equation}
here $\textbf{W}^s \in \mathbb{R}^{d \times s}$ is the filter weight of window size $s$. $ \textbf{x}^c_{t:t+s-1} $ consists of $s$ character embeddings starting from the location $t$. Then we apply max-pooling across different n-grams convolution results at each location $t$:
\begin{equation}
		 \textbf{x}^p_t = max(\hat{\textbf{x}}^p_{1,t}, \hat{\textbf{x}}^p_{2,t}, \hat{\textbf{x}}^p_{3,t}) \in \mathbb{R}^{d}. \qquad  t \in \{1,2,...,T\} 
\end{equation}
This 1-D convolution and max-pooling learn to adaptively select different gram features at each location and preserve the original sequence length and order. After that, we obtain the phrase vectors $\textbf{X}^p = [\textbf{x}^p_1, .., \textbf{x}^p_T] \in \mathbb{R}^{d \times T}$. 

\textbf{Multiple Attention}. For the decoder part, we employ GRU RNN with attention mechanism \cite{bahdanau2014neural} to generate the next line. Here we take $\textbf{X}^c $, $\textbf{X}^p $, and $\textbf{X}^s $ as three kinds of hierarchy attention memories and calculate attended context vectors. Since the dimension of $\textbf{x}^s_t$ is twice of $\textbf{x}^c_t$ and $\textbf{x}^p_t$. In order to make these dimensions equal, we design two variants: 1) We concatenate $\textbf{x}^c_t$ and $\textbf{x}^p_t$ for each time step $t$ and obtain $\widetilde{\textbf{X}}^{cp} = [\widetilde{\textbf{x}}^{cp}_1, .., \widetilde{\textbf{x}}^{cp}_T] \in \mathbb{R}^{2d \times T}$. Then we concatenate $\textbf{X}^s $ and $\widetilde{\textbf{X}}^{cp}$ across time step as the hierarchy attention memory $\textbf{H}^{concat} \in \mathbb{R}^{2d \times 2T}$. 2) We tile each $\textbf{x}^c_t$ twice individually and obtain $\widetilde{\textbf{X}}^c = [\widetilde{\textbf{x}}^c_1, .., \widetilde{\textbf{x}}^c_T] \in \mathbb{R}^{2d \times T}$. We do the same for $\textbf{x}^p_t$ to obtain $\widetilde{\textbf{X}}^p = [\widetilde{\textbf{x}}^p_1, .., \widetilde{\textbf{x}}^p_T] \in \mathbb{R}^{2d \times T}$. Then we concatenate $\textbf{X}^s $, $\widetilde{\textbf{X}}^c$, and $\widetilde{\textbf{X}}^p$ across time step as the hierarchy attention memory $\textbf{H}^{tile} \in \mathbb{R}^{2d \times 3T}$. 

The $i$-th GRU hidden state \textbf{s}$_i$ of decoder part is calculated as:
\begin{equation}
	\textbf{s}_i = \textbf{GRU}(\textbf{g}_i,\textbf{s}_{i-1}).
\end{equation} 
Here \textbf{g}$_i$ is linear combination of attended context vector \textbf{c}$_i$ and the character embedding of (i-1)-th character $ \textbf{y}_{i-1}$:
\begin{equation}
	\textbf{g}_i = \textbf{W}^y\textbf{y}_{i-1}+ \textbf{W}^c\textbf{c}_i. 
\end{equation} 
The attended context vector \textbf{c}$_i$ is computed as a weighted sum of the hierarchy attention memory $\textbf{H}$:
\begin{equation}
	\textbf{c}_i = \sum_j{\alpha_{ij}\textbf{H}_j}.
\end{equation} 
And the equation for calculating the weight $\alpha_{ij}$ of each \textbf{H}$_j$ is as follows:
\begin{equation}
	\alpha_{ij} = \frac{exp(e_{ij})}{\sum_k{exp(e_{ik})}}.
\end{equation} 
Where
\begin{equation}
	e_{ij} = \textbf{v}_a^T \textbf{tanh}(\textbf{W}^a\textbf{s}_{i-1}+\textbf{U}^a\textbf{H}_j).
\end{equation} 
We define each conditional probability as:
\begin{equation}
	P(y_i | S^m_{1:i-1},S^1,...,S^{m-1}) = \textbf{Softmax}(\textbf{W}^o\textbf{s}_i+b).
\end{equation}

\textbf{Reranking}. Given previous $m-1$ generated lines $\{S^1, ..., S^{m-1}\}$, we implement beam search to generate $k$ candidate $m$-th lines $\{S^{m}_1,...,S^{m}_{k}\}$. Here $k$ is the beam width. To make the whole poem more relevant with the title, we rerank all candidate lines by the pre-defined score. The score of $j$-th candidate sentence of $m$-th line $S^m_j$ is defined as:
\begin{equation}	
	\textbf{score}(S^m_j) = (100-\textbf{PPL}(S^m_j)) \times \max_{t'_k\in S^m_j}{\phi(\textbf{T}(t'_k), \textbf{T}(t^*))}.
\end{equation} 
Here $t^{*}$ is the title and $ S^m_j $ is segmented into a set of phrases $ \{t'_1, \cdots, t'_k\} $. The second term of above equation measures the correlation between the sentence and the title. The PPL in the first term is the perplexity \cite{jelinek1977perplexity} which is one kind of important metric of Nature Language Processing (NLP). The PPL of sentence $S$ is defined as follows:
\begin{equation}
	  	 \textbf{PPL}(S) \equiv 2^{-\frac{1}{n}\sum \limits_{i=1}^n \log{P(w_i | w_1,...,w_{i-1})}},
\end{equation}
where $n$ is the length of sentence $S$ and $w_i$ is the $i$-th token.

We take the highest score candidate sentence as the $m$-th line of the generated poem.

\section{Experiments}
Our experiments revolve around the following questions: \textbf{Q1:} As we introduce the phrase feature into the \textit{HieAS2S} model, does this feature help? Which configuration is the most effective one? \textbf{Q2:} Judging from the human views, how does the proposed three-stage approach compare with the strong baseline? \textbf{Q3:} Whether our method can generate the suitable titles for the generated poems?

\subsection{Dataset}
We built a large poetry corpus called \textit{corpus-P} which contains 149,524 traditional Chinese poems in various genres. The most poems in \textit{corpus-P} are \textit{quatrains} or \textit{regulated verses}. We randomly chose 3000 \textit{quatrains} for validation and 3000 \textit{quatrains} for testing. After the preprocessing of low frequency characters, the size of vocabulary is 5295. This poetry corpus was used to train B/F-LM and HieAS2S model. Another external large corpus (\textit{corpus-M}) including 18,657 Chinese Song Iambics, 17,000K characters from ancient Chinese proses and poetry corpus were used to train LDA model and pre-train character embeddings. For \textit{image recognition module}, we manually filtered and clustered themes of \textit{ShiXueHanYing} into 40 classes, and built a image dataset including over 40,000 pictures. 1000 pictures were randomly chosen for each class as train set and 100 for each as test set.

\begin{figure*}[htbp]
 \label{show}
  \centering
  \subfigure[The outputs of \textit{image recognition module} for the user-uploaded image.]{
  \begin{minipage}{5.0cm}
  \centering
    \includegraphics[width=5.0cm]{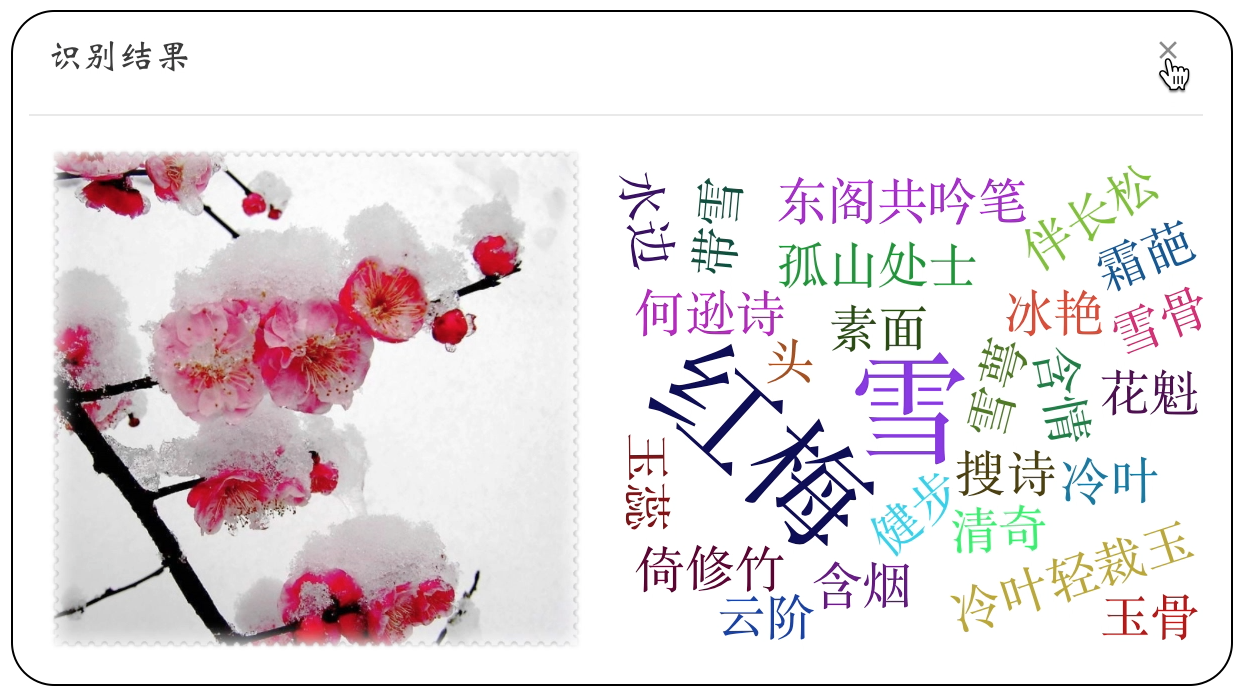}
  \end{minipage}
  }
  \subfigure[A 5-character \textit{quatrains} generated with the user-uploaded image.]{
  \begin{minipage}{5.0cm}
  \centering
    \includegraphics[width=5.0cm]{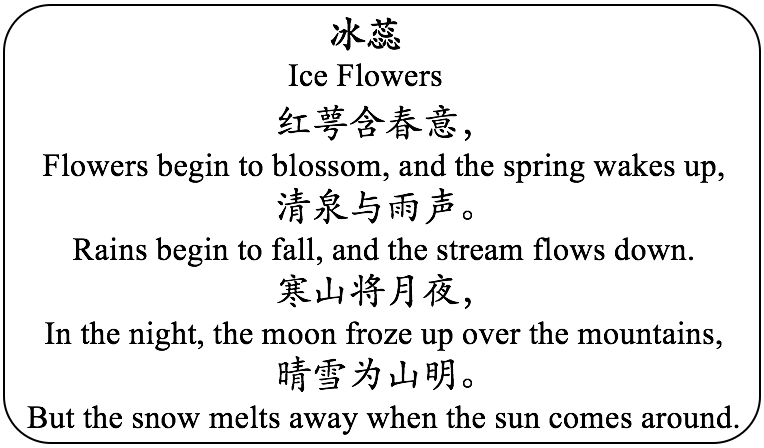}
  \end{minipage}
  }
  \subfigure[An example of 7-character \textit{quatrains} which is generated with a given theme `loneliness'.]{
  \begin{minipage}{5.0cm}
  \centering
    \includegraphics[width=5.0cm]{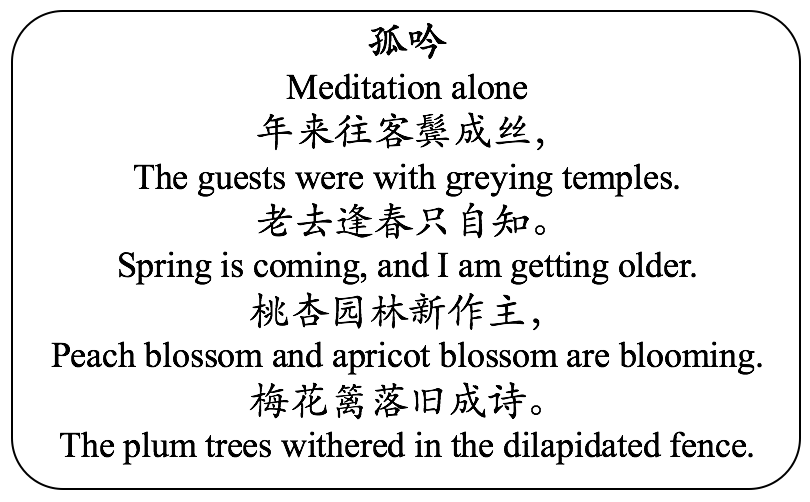}
   
  \end{minipage}
  }
  
  \caption{Figure (a) shows the outputs of \textit{image recognition module} of the user-uploaded image. We visualize the theme of the image ``red plum" and other theme related phrases of \textit{Shixuehanying} to users. Figure (b) shows a 5-character \textit{quatrains} generated with the image of ``red plum". Figure (c) shows an example of 7-character \textit{quatrains} which is generated with a given theme `loneliness'.}

 \end{figure*}
 
\subsection{Training}
For LDA model training, the Jieba Chinese text segmentation\footnote{https://github.com/fxsjy/jieba} (a Python based Chinese word segmentation module) is employed for segmentation and building dictionary for LDA model. Particularly, we added all theme related phrases of \textit{ShiXueHanYing} to this dictionary. After that, we used gensim\footnote{https://radimrehurek.com/gensim/} (a free Python library for NLP) to help us to train a 100-topics LDA model on the \textit{corpus-M}. The experiments indicate that using \textit{corpus-M} instead of \textit{corpus-P} to train LDA model is conducive to improve the performance of LDA model, in terms of PPL.

We used noise-contrastive estimation (NCE) \cite{mnih2013learning} method to pre-train 512-dimension character embeddings with a skip-gram model \cite{mikolov2013distributed}. Priori knowledge was brought into the model by these character embeddings which were trained on \textit{corpus-M}. Note that some characters of \textit{ShiXueHanYing} rarely appear in \textit{corpus-P}, the pre-trained character embeddings can help the models to learn them better. 

To train the B/F-LM model, we further pre-processed the training data. For each poem in the training data, if the first line of the poem contains a phrase in \textit{ShiXueHanYing}, we used it as the split word and reversed the first half of the line to train the backward LM. Otherwise, we randomly picked one word as the split word. For the \textit{HieAS2S} model training, we followed \cite{Wang2016Can} and used their proposed training strategy called hybrid-style training (training 5-char poems and 7-char poems using the same model with a type indicator) to improve the model. We used Adam optimization method \cite{kingma2014adam} with 128 mini-batch size for training. The learning rate was set to 0.001 which can be a constant or dynamically set \cite{lv2007convergence,lv2009convergence}. Several techniques were investigated to train and improve the model, including RNN-dropout \cite{Gal2015A}, gradient clip and weight decay. The hyper-parameters were chosen empirically and adjusted in the validation set. It is worth mentioning that we found our models equipped with GRU cells performed slightly better than LSTM cells in our experiments.

\subsection{The Ablation Study (Q1)} 
In the first experiment, we aimed to test the effectiveness of the proposed model. We evaluated and compared the \textit{HieAS2S} model with several variants. Four different models were tested, they were: 

1) The standard seq2seq model with attention mechanism (\textit{baseline}). The attention memory of this model is $\textbf{X}^s \in \mathbb{R}^{2d \times T}$. 

2) The seq2seq model whose attention memory consists of $\textbf{X}^s \in \mathbb{R}^{2d \times T}$ and $\textbf{X}^c \in \mathbb{R}^{d \times T}$. This model is presented by \cite{Wang2016Can} called \textit{AS2S}. 

3) The proposed \textit{HieAS2S} model whose attention memory is $\textbf{H}^{concat} \in \mathbb{R}^{2d \times 2T}$ called \textit{HieAS2S-concat}.

4) The proposed \textit{HieAS2S} model whose attention memory is $\textbf{H}^{tile} \in \mathbb{R}^{2d \times 3T}$, we called it \textit{HieAS2S-tile}. 

To be fairness and reduce the impact of the first line generation, here we didn't implement our first line generation method to generate first lines. Instead, we randomly picked 1000 poems from the test set and took their first lines as inputs for above models to generate the whole poems.

For evaluation, referring to \cite{Wang2016Can, zhang2014chinese, Yi2017Generating}, we used BLEU-2 score as a cheap evaluation metric to evaluate these 4000 generated poems. Here we slightly modified this method. Since each poem was generated by a given first line, we constructed the reference set as follows: For each \textit{ShiXueHanYing} theme, we firstly counted the number of co-occurrence words for each poem in the dataset and the related phrases of the theme. We retained the top 20 poems with the largest number of co-occurrence words as the reference set for each theme. We used the same method to judge the themes of the poems whose first lines were used for poetry generation. Then for each generated poem, we retrieved the theme of the original poem of its first line, and took the reference set of this theme as the reference set of this generated poem. In order to ensure the effectiveness of this modified BLEU method, we did a comparative experiment with positive and negative examples. For these 1000 poems whose first lines were used to generate poetry, we firstly calculated their BLEU scores called the \textit{positive-groundtruth} scores with their themes' reference sets. Then for each of these poems, we replaced its theme with a random \textit{ShiXueHanYing} theme and calculated its BLEU score as the \textit{negative-groundtruth} score. 

\begin{table}[t]
\renewcommand{\arraystretch}{1.3}
\caption{The BLEU-2 scores and RHYTHM scores of different approaches}
\label{BLEU}
\centering
\begin{tabular}{|c|c|c|}
 \hline
 \multirow{2}{*}{Approach} &
  \multicolumn{2}{c|}{Metrics} \\
 \cline{2-3}
  & BLEU-2 &  RHYTHM  \\
 \hline 
\textit{baseline} & 26.726 & 0.824 \\ 
 \hline
 \textit{AS2S} & 27.458 & 0.866  \\ 
 \hline
 \textit{HieAS2S-tile} & \textbf{29.991} & \textbf{0.892}  \\ 
 \hline
 \textit{HieAS2S-concat} & 28.171 & 0.876  \\ 
 \hline
 \textit{Positive-groundtruth} & 29.095 & 1.000 \\ 
 \hline
 \textit{Negative-groundtruth} & 12.062 & 1.000  \\ 
 \hline
 \end{tabular}
\end{table}

Because the above BLEU method can't evaluate the rule-consistency of generated poems, we followed \cite{Yang2017Generating} and used the RHYTHM score for further evaluation. The RHYTHM score is used to measure whether a generated poem meet the constraints of tonal and rhyme which is defined as follows:
\begin{equation} 
	\begin{split} 
	  	 & \textbf{RHYTHM}(l) =    		
		 \begin{cases}
		 	0, &\mbox{$cnt(l) \notin \{5, 7\}$}\\
   			0.5, &\mbox{$ rule(l) \in T \  or \  R$}\\
   			1.0, &\mbox{$rule(l) \in T \  and \ R$}
  		 \end{cases} 
	\end{split} 
\end{equation}
where $l$ denotes a poem line, $cnt(l)$ denotes the length of $l$, and $rule(l) \in T$ means $l$ meets the constraints of tonal while $rule(l) \in R$ means meets the constraints of rhyme. 

The results are shown in Table \ref{BLEU}. From the last two rows of the table, we can see that the BLEU-2 scores of \textit{positive-groundtruth} is much higher than \textit{negative-groundtruth} which shows that the modified BLEU score is effective. As we can see from the first four rows of the table, the \textit{AS2S} performs better than \textit{baseline}. And both of the proposed model \textit{HieAS2S-tile} and \textit{HieAS2S-concat} outperform other models, in both terms of BLEU-2 scores and RHYTHM scores. In addition, the \textit{HieAS2S-tile} model performs better than \textit{HieAS2S-concat}. Through our analysis, the \textit{HieAS2S-tile} model divides word features and phrase features separately, so that the model can better capture the phrase information. This results show that the phrase features are helpful and demonstrate the effectiveness of our proposed models.

\subsection{Human evaluation (Q2)}
In the second experiment, we compared the proposed three-stage approach with the strong baseline \textit{AS2S} \cite{Wang2016Can} by human evaluation. In this second experiment, we used \textit{HieAS2S-tile} instead of \textit{HieAS2S-concat}. Since human evaluation is time-consuming and laborious, we mainly compared the proposed method with one of the most popular approaches which achieved the state-of-the-art performance to reduce human efforts. The \textit{AS2S} model was fully compared with most of the previous poetry generation approaches such as SMT \cite{he2012generating}, Seq2Seq \cite{sutskever2014sequence}, LSTM language model \cite{sundermeyer2012lstm}, and RNNPG \cite{zhang2014chinese} in \cite{Wang2016Can}. It has shown that this approach performs better than the rest of the approaches, so we didn't compare our method with those approaches. In addition, both our proposed method and \textit{AS2S} can generate poetry by given \textit{ShiXueHanYing} themes. 

For each method, we selected 30 \textit{ShiXueHanYing} themes to generate 60 quatrains with beam size 10. For further comparison, we also involved 40 unfamous human-written quatrains in the evaluation. We invited 10 human experts to evaluate these 160 poems. Following \cite{he2012generating, zhang2014chinese, Wang2016Planning}, we set four evaluation standards for human evaluators to judge the poems: \textbf{Poeticness}, \textbf{Fluency}, \textbf{Coherence} and \textbf{Meaning}. The score of each aspect ranges from 1 to 5 with the higher score the better. The detailed illustration is listed below:

(a) \textbf{Poeticness}: Does the poem follow the rhyme and tone requirements?

(b) \textbf{Fluency}: Does the poem read smoothly and fluently?

(c) \textbf{Meaning}: Does the poem have a certain meaning and artistic conception?

(d) \textbf{Coherence}: Is the poem coherent across lines?

Table \ref{human} presents the results. Our method performs better than AS2S in all four metrics. This results show the effectiveness of our method. Compared our method with human-written, we found that the \textbf{Poeticness} and \textbf{Fluency} scores of our method are slightly lower than human-written poems. However, the \textbf{Meaning} and \textbf{Coherence} scores of poems written by human are still much higher than those generated by our method. Particularly, we developed a web application for users to use and evaluate our approach, most users give satisfactory evaluations to our approach. Figure 3 shows an example of 5-char quatrain generated on our web application with a user-uploaded picture and another 7-char quatrain generated with a given theme.

\begin{table}[t]
\renewcommand{\arraystretch}{1.3}
\caption{The results of human evaluation}
\label{human}
\centering
\begin{tabular}{|c|c|c|c|c|c|}
 \hline
 Method & Poeticness & Fluency & Meaning & Coherence & Overall \\ 
 \hline 
AS2S & 3.62 & 3.07 & 2.73 & 3.12 & 3.13 \\ 
 \hline
Ours & 3.87 & 3.24 & 2.85 & 3.24 & 3.30 \\ 
 \hline
Human-written &  4.07 & 3.43 & 3.58 & 3.71 & 3.69 \\ 
 \hline
\end{tabular}
\end{table}
 
\subsection{Title evaluation (Q3)}
In this experiment, we evaluated our title generation method. We compared the proposed title generated approach with standard \textit{Seq2Seq} model which has achieved good results on abstractive summarization \cite{Rush2015A,Chopra2016Abstractive}. We filtered the poems whose titles are longer than 15 characters or contain low-frequency characters on \textit{corpus-P}. After that, these poem-title training pairs were used to train a GRU seq2seq model with attention mechanism.

For evaluation, we firstly implemented our three stage methods to generate 100 poems (including titles) with random \textit{ShiXueHanYing} themes. Secondly, these 100 poems (without the titles) were fed into the seq2seq title generation model to generate another 100 titles. Finally we did a pair comparison experiment. Given each generated poem and its two different titles generated by two methods, we asked the experts to decide which title is more appropriate. 

\begin{figure}[!t] 
   \centering
   \includegraphics[width=2.5in]{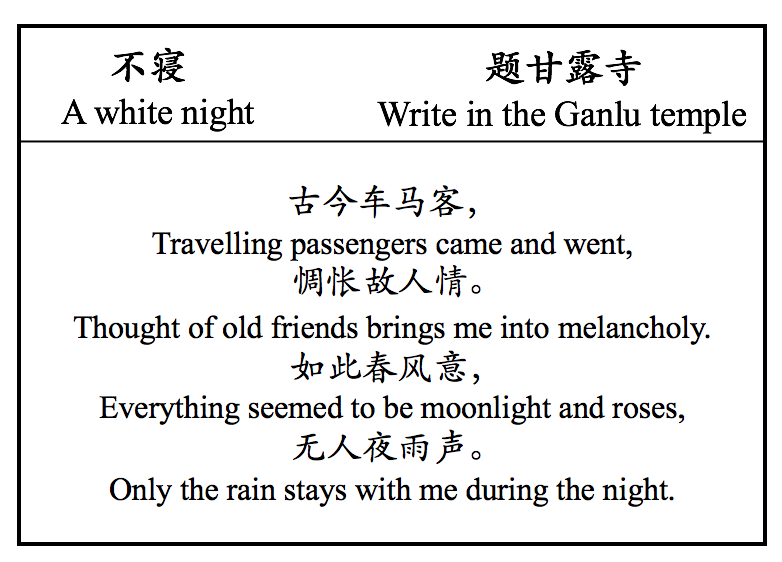}
   \caption{An example of the title evaluation experiment which contains a poem and a pairs of titles. The title ``The white night'' on the left hand side is generated by our methods while another title ``Write in the Ganlu temple'' is generated by the seq2seq model. The experts prefer the previous title, because the second title doesn't seems to be related to the poetry.}
   \label{title}
\end{figure}

The result of our method \textit{vs} \textit{Seq2Seq} is 83:17. This result shows that our method significantly outperforms the \textit{Seq2Seq} model. We found \textit{Seq2Seq} model tends to generate very general titles such as ``Early Spring", ``Send to friend", and ``Departure". In addition, the \textit{Seq2Seq} model also often generate some titles which contain specific geographical or landscape names which are not related to poetry. We show an example of the test which contains a poem and a pairs of titles in Figure \ref{title}. 

\section{Conclusion and Future Work}
In this paper, we have three contributions:
\begin{itemize}
\item We propose a three-stage approach to generate Chinese \textit{quatrains}. This approach is able to generate high-quality \textit{quatrains} with relevant titles. Our experiments demonstrate that the proposed methods of title generation and poetry generation both outperform the strong baselines.

\item We introduce a multi-modal way for Chinese \textit{quatrains} generation. We extend the generation way to support poetry generation through pictures. Furthermore, we manually build a large image-to-theme dataset.

\item We add phrase feature to poetry generation model and propose the \textit{HieAS2S} model. Our experiments show that this phrase information is helpful and the \textit{HieAS2S} model performs better than several variants and strong baseline.
\end{itemize}

In the future, we will explore the following further work: 
\begin{itemize}
\item Extending our approach to generate Song Iambics which have bigger challenges. 

\item Focusing on combining semantic image segmentation to further strengthen the relationship between images and poetry.
\end{itemize}

\section*{Acknowledgment}
This work was supported by the National Science Foundation of China (Grant No.61625204), partially supported by the State Key Program of National Science Foundation of China (Grant No.61432012 and 61432014).

\bibliographystyle{IEEEtran}

\bibliography{PoetryBib}

\end{document}